\documentclass[lettersize,journal]{IEEEtran}
\usepackage{amsmath,amsfonts}
\usepackage{algorithmic}
\usepackage{array}
\usepackage[caption=false,font=normalsize,labelfont=sf,textfont=sf]{subfig}
\usepackage{textcomp}
\usepackage{stfloats}
\usepackage{url}
\usepackage{verbatim}
\usepackage{graphicx}
\usepackage{multicol}
\usepackage{lipsum}
\def\BibTeX{{\rm B\kern-.05em{\sc i\kern-.025em b}\kern-.08em
    T\kern-.1667em\lower.7ex\hbox{E}\kern-.125emX}}
\usepackage{balance}
\begin{document}
\title{GC-GRU-N for Traffic Prediction using Loop Detector Data\\
}
\author{\textsf{Maged Shoman,  Armstrong Aboah, Abdulateef Daud \& Yaw Adu-Gyamfi}
	
	Department of Civil and Environmental Engineering
	University of Missouri
	Columbia, MO, USA
	
	\texttt{mas5nh, aa5mv, aadcvg, adugyamfiy@missouri.edu}
}

\maketitle

\begin{abstract}
Because traffic characteristics display stochastic nonlinear spatiotemporal dependencies, traffic prediction is a challenging task. In this paper develop a graph convolution – gated recurrent unit (GC-GRU-N) network to extract the essential spatio-temporal features. we use Seattle loop detector data aggregated over 15 minutes and reframe the problem through space and time. The model performance is  compared o benchmark models; Historical Average, Long - Short Term Memory (LSTM) and Transformers. The proposed model ranked second with the fastest inference time and a very close performance to first place (Transformers). Our model also achieves a running time that is six-times faster than transformers. Finally, we present a comparative study of our model and the available benchmarks using metrics such as training time, inference time, MAPE, MAE and RMSE. Spatial and temporal aspects are also analyzed for each of the trained models. 
\end{abstract}

\section{Introduction}
The path toward an Intelligent transportation system (ITS) has recently become more feasible due to the influence of two major factors: 1) exponential growth of data collected by embedded traffic sensors, and 2) advancement of effective deep learning techniques [1]. An intelligent transportation system consists of several components, one of which is traffic forecasting. Traffic forecasting is an essential component of an ITS because it predicts future traffic flows on road networks by analyzing both historical traffic data and the configuration of road networks. Forecasted traffic flows are required for several traffic management applications, including traffic control [2, 3], traffic classification [4, 5], and vehicle scheduling [6]. Despite its many benefits, traffic forecasting still remains a daunting task. 
Traffic forecasting is a challenging task [7-23] because traffic variables such as speed, volume, and traffic patterns are influenced by dynamic and static factors known as spatiotemporal correlations and external events. The above-mentioned factors can profoundly influence the performance of a traffic forecasting system directly and indirectly. First, studies have shown that spatial information, precisely the locations of embedded communication sensors, significantly influences a traffic forecasting system [7]. This is because roads in a Euclidean space are bound to have different traffic conditions at any given time [7]. For example, on a two-lane highway network, there is usually a significant difference in the amount of traffic traveling in each lane at any given time. Also, the traffic speed on a given roadway is influenced and directed by the traffic condition further downstream. Second, traffic dynamics and their temporal dependencies can be different from one another by combining recurring patterns and unpredictability of occurrences. For example, traffic follows a cyclical pattern on a daily and weekly basis, but there may be dynamic shifts in temporal patterns due to crashes, and this results in difficulties during traffic forecasting. Third, extraneous factors such as one-time events and weather conditions can significantly impact traffic flow, making long-term traffic forecasting more difficult. 

Several methodologies for short- and long-term traffic forecasting have been implemented in light of the numerous challenges identified in forecasting traffic. First, models such as K-nearest neighbor (KNN) [8,9], Gaussian process [10], hidden Markov model [10], support vector regression (SVR) [11], and autoregressive integrated moving average (ARIMA) [12,13] were used in the past. Typically, these techniques are limited to less complex traffic conditions and situations with small data. Second, deep-learning-based methods, primarily recurrent neural networks (RNNs) and convolutional neural networks (CNNs). RNNs, such as long-short-term memory (LSTM) [14] and gated recurrent unit (GRU) [15], are commonly used for sequential and temporal learning, whereas CNNs, such as ResNet [16], are commonly used for learning spatial structures. ST-ResNet [17] is a time-series model that uses a residual network and LSTM. Although the aforementioned approaches have produced cutting-edge results, they do not consider the connectivity of road networks. This is significant because traffic conditions on one road will be influenced by another. As such, methodologies need to consider both traffic variables and the road network's configuration.
Most recently, researchers have begun modeling traffic data collected by road sensors using graph-theoretic approaches. The spatial correlations between traffic sensors are represented by a directed graph with nodes representing the sensors and edge weights representing the proximity of sensor pairs as determined by road network distance. Recent advancements in graph neural networks [20], especially convolutional graph neural networks [18], have fueled the development of several graph-based traffic prediction models [19-21] because sensor networks are naturally organized as graphs, as in the case of DCRNN. DCRNN [7] represents the road network as a directed weighted graph and proposes a diffusion convolutional RNN for traffic prediction. Even though GCN has achieved great success in spatial analysis over the years, few studies have looked into using GCN for spatiotemporal analysis. [31] combined recurrent neural networks and GCN to perform spatiotemporal analysis. Although a significant result was obtained, the architecture is constrained by the limitations of the RNN, which cannot be used for long-term forecasting due to vanishing and exploding gradients. Other studies, such as [46], investigated combining LSTM and GCN to capitalize on LSTM's ability to learn from long-term dependencies. This was a significant accomplishment, as the results of these studies achieved state-of-the-art. Even though LSTM has achieved respectable results in recent years, training LSTM takes longer [40]. Alternatively, studies have investigated the performance of Gated Recurrent Unit (GRU) for long sequence prediction and concluded that its results are comparable to that of the LSTM. The GRU has the advantage over the LSTM because it is less complex and can be trained more quickly. 

The primary goal of this paper is to develop a graph convolution – gated recurrent unit (GC-GRU-N) model to perform network–wide traffic forecasting. The dataset for this study was collected from the TRBAIAC traffic4cast challenge [54] as inductive loop detectors installed on freeways throughout the Greater Seattle area. This dataset contains freeway traffic performance score (TPS), speed, and volume information. The freeways include I-5, I-405, I-90, and SR-520. The traffic states of loop detectors on main lanes traveling in the same direction are aggregated every two miles in this dataset. In our proposed model, the GRU cells was used to model the temporal aspect of the problem, while the GC cells were used to model the spatial aspect of the problem based on the road network configuration using the adjacency matrix. 
The rest of the paper is structured as follows. Section two examines previous traffic forecasting methodologies. The third section explains the data used in this paper. Section four goes over the methodology used in this study. Section five presents and discusses the results of our study. Sections six and seven present the study's conclusion and recommendations.

\section{Literature Review}
\noindent This section reviews methodologies employed extensively by previous studies. Numerous traffic network modeling methodologies have been identified as useful for estimating and predicting traffic patterns. In the past, parametric statistical models were utilized frequently to model traffic data flow. Several studies favored the Autoregressive Integrated Moving Average Model (ARIMA) because of the model's capacity to model sequential input [24–26]. 
In addition, various approaches based on machine learning have been suggested as possible ways to model traffic data. In previous studies, researchers made use of techniques such as Support Vector Regression (SVR) [27], [33], [34], k-nearest neighbors (k-NN) [35], [36], Bayesian Networks (BN) [37], and feed-forward Neural Networks (NN) [28], [29], [38], [39]. A new frontier has been opened for traffic modeling based on deep learning due to traffic data's growing pervasiveness, availability, and size. Deep neural network–based methods have recently been shown to achieve high accuracies for traffic estimation and prediction tasks due to the availability of large amounts of data pertaining to traffic. This was made possible by the availability of big data. RNNs, or recurrent neural networks, are a subcategory of deep neural networks developed specifically to model sequential data. Long Short-Term Memory (LSTM) and Gated Recurrent Units (GRUs), which are both subclasses of RNNs, were used extensively in recent research to model the spatiotemporal behavior of traffic [7], [40]– [41]. For instance, Cui et al. [42] proposed a stacked bidirectional and unidirectional LSTM network (SBU-LSTM) to model traffic's chronological and reverse chronological temporal dependencies. Many authors used convolutional neural networks to improve the spatial modeling capability of deep learning models. This was done by combining multiple layers of neural connections (CNNs). 
CNNs can model both local and global relationships between neighboring pixels as they were initially developed for use in computer vision applications. CNNs were utilized in a great number of research projects to model the connection between neighboring stretches of roadway [43–44]. For instance, Ke et al. [48] modeled lane-wise traffic speed and volume data by employing a CNN with multiple channels to analyze the data. The proposed method was developed to capture the spatial relationship between the traffic lanes that are immediately adjacent to one another. Graph Convolutional Networks (GCN) were also used to model spatial traffic dependencies [30–32], [45], and [46]. These dependencies were derived from the topological structure of the road network. In their study [31], Zhao et al. combined GCN for spatial modeling with GRU for temporal modeling, and the result was a Temporal Graph Convolutional Network (T-GCN). T-GCN was trained and tested for its ability to predict traffic speed using two distinct datasets: data based on a probe-vehicle type and data based on a loop detector. An Attention-based Spatio-Temporal Graph convolutional network (ASTGCN) was proposed in a paper by Guo et al. [47]. To make an accurate traffic flow prediction, the authors utilized data on traffic flow, occupancy, and speed collected from sensors embedded in the infrastructure.

\section{Data}
\noindent The importance of Traffic Performance Measurement cannot be overemphasized. This is because it offers a bi-folded benefit to both the Transportation Agencies who use the metric to make data-driven policies for effective infrastructural operation and to the general public to plan their trip efficiently. Most of the existing performance measures are either road segment or trip-based, they fail to evaluate the road performance over a complicated network. To this end, [52] at the University of Washington’s STAR Lab propounded the Traffic Performance Score (TPS) which is a multiple-parameter indicator for accessing traffic performance at the network level.
Ideally, the physical properties of road segments are fixed, however the traffic stream parameters on each segments including Volume(Q), Speed(V), and density(K) keeps changing. In order to evaluate the network’s traffic performance, the parameters of each road need to be factored in.  The TPS for a segment is calculated as:
\begin{figure}[!t]
	\centering
	\includegraphics[width=3.5in]{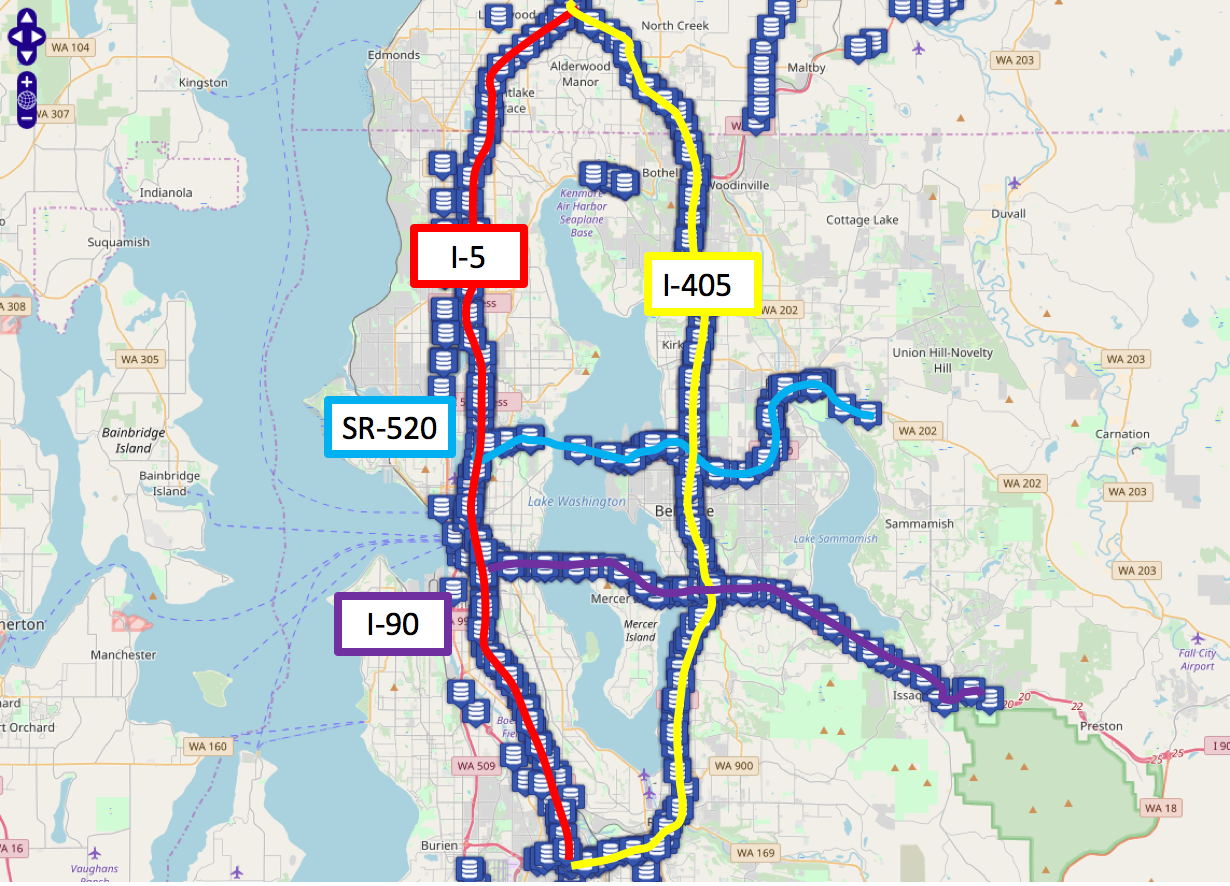}
	\caption{Freeways in study area}
	\label{fig1}
\end{figure}

\begin{equation}
	\label{deqn_ex1}
	TPS^t= \frac{\sum_{i=1}^{n} V_i^t Q_i^t L_i}{\sum_{i=1}^{n} V_f Q_i^t L_i}
\end{equation}

where $i, n, t, v, v_f, Q,$ and $L$ represent segment loop detector index, number of segments, time, speed, free-flow Speed, volume and length respectively. TPS is a value ranging from 0\% to 100\%. Overall network-wide traffic condition is best when TPS is 100\% and the worst when TPS is 0\%

The data used for the estimation of Traffic Performance Score is calculated based on the data collected from almost 8000 inductive lop detectors deployed on the freeway network in the Northwest region in Washington State [54]. The freeways covered in the study includes I-5, I-405, I-90, and SR-520 as shown in Fig. 1. The dataset consists of traffic performance score, spatio-temporal speed and volume information of the freeway system. Each blue icon represents the loop detectors at a milepost, the speed data at a milepost is averaged from the several loop detectors on the main lanes in the same direction at a particular milepost.
The training dataset used in the study is sampled from January 1st, 2020, to May 31st, 2020, at 15-minute sampling interval.

In order to test the performance of the model, there are 15 testing data points which will represent the ground truth in the study with 36 previous time steps.  The test data covered weekday, weekends, morning and afternoon peak hours. The developed model is required to predict the next 12 steps ahead of ‘TrafficIndex\_GP’.
 
\section{Methodology}
\noindent In this section, we discuss our end-to-end traffic prediction model, designed to tackle the traffic forecasting challenge using the loop detectors data, which consists of spatial-temporal GC-GRU cells for the encoder and decoder components. The overall architecture of the proposed model is illustrated in Fig 2. Specifically, the proposed architecture consists of two main parts: encoder and decoder. The encoder module consists of GC and GRU cells to encode the traffic features (input) and the decoder module consists of GC and GRU cells to forecast traffic parameters (output) from the encoded state. GC is employed to learn the complex spatial relationships between traffic detectors and GRU is utilized to capture the dynamic temporal dependencies of traffic data reported by traffic detectors at different times.

\begin{figure}[!t]
	\centering
	\includegraphics[width=3.6in]{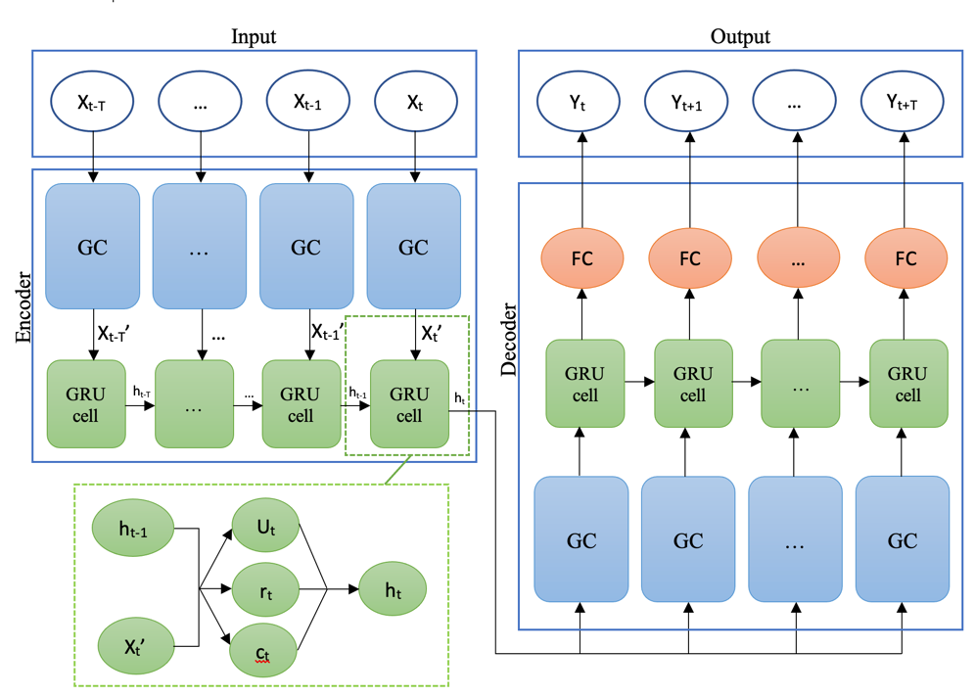}
	\caption{GC-GRU-N proposed architecture}
	\label{fig2}
\end{figure}

\subsection{GC for spatial relationships}
\noindent In the past decade, neural networks have experienced tremendous success. Although many data sets in the real world contain underlying graph structures that are not Euclidean, early neural network variations could only be built using regular or Euclidean data. New developments in Graph Neural Network (GNN) have been made possible by the non-regularity of data structures. GCN are one of the many variations of GNNs that have been developed over the last several years and are regarded as one of the fundamental Graph Neural Networks subtypes. The actions are carried out by GCNs are similar to Convolutional Neural Network (CNN), however the model learns the features by looking at the nearby nodes. The primary distinction between CNNs and GNNs is that the former was developed specifically to function on regular (Euclidean) organized data, whilst the latter are generalized CNNs with varying numbers of connections and unordered nodes.
In our proposed model, the input to the model is a network of traffic loop detectors represented by nodes (\text{$N$}) and edges (\text{$E$}). Nodes represent the location or order of the detectors while edges represent the existence of a link in between the nodes. This can be represented by an undirected graph where \text{$G = (N , E)$} with \text{$N$} defined as the set of detectors. 
One detector \text{$ni$} is connected by vertices \text{$i$} and \text{$j$} which represent the network edges (\text{$ni , nj$}). 
To consider the impact of traffic events, bidirectionally on upstream and downstream roads, we let \text{$G$} be an undirected graph even if certain roads are directed.              
Subsequently, the Adjacency square matrix (\text{$A$}) element for each \text{$E$} is represented as  \text{$Ai,j = Aj,j$} with values of {0,1}, in which \text{$Ai,j = 1$} for an existing connection between nodes \text{$ni$} and \text{$nj$} and \text{$Ai,j$} = 0 otherwise. Node self-features are also considered by performing a self-loop, adding the adjacency and identity matrix using the following formula for the -hop neighborhood of a node (clipping nodes that might exceed one): 

\begin{equation}
	\tilde{A}_{i,j}^k = min((A+1)_{i,j}^k,1)
\end{equation}

To capture the spatial properties in the coded graph (\text{$G$}), the GC layer builds a filter in the Fourier domain that acts on the nodes and their first-order neighbors using the input adjacency matrix (\text{$A$}). Adjacency matrix weights for all edges are then input into a diagonal vector (\text{$D$}) that now contains information about the degree (edges count) of each vertex. The stronger a node's affiliation with a particular group or cluster, the lower its degree. To ensure numerical stability during training, symmetric normalization is then performed using the following formula: 
\begin{equation}
	\hat{A}_{i,j}^k = \tilde{D}_{i,j}^k{-\frac{1}{2}}\tilde{A}_{i,j}^k\hat{D}_{i,j}^k{-\frac{1}{2}}
\end{equation}

Output for each GC layer (\text{$z$}) is then calculated using the dot product of the normalized adjacency matrix (\text{$\hat{A}_{i,j}^{k} $}) , input feature (\text{${X}_{i,j}^k $}) and the weight matrix (\text{${W}_{i,j}^k{i-1}$}) of the previous GC layer. Rectified Linear Activation Unit ({$ReLu$}) is then used as the activation function. The calculation can be expressed using: 
\begin{equation}
    L^z = ReLu(\hat{A}_{i,j}^k{X}_{i,j}^k{W}_{i,j}^k{}^{(i-1)})
\end{equation}
The output from our 2-layer GC model can finally be expressed using the following: 
\begin{equation}
    \label{eqdfkj2etft22}
    gc(S_t, A) = \hat{A}_{i,j}^k {L}_{i} {W}_{i,j}^k{}^{(i)}
\end{equation}

\subsection{GRU for temporal  relationships}
\noindent Another significant issue with traffic prediction is acquiring temporal dependence. Given the exceptional time series predictions, the RNN has shown good results over the past decade. However, because of flaws like gradient disappearance and explosion, classical RNN has limitations for long-term projections. To solve these issues, the LSTM cell and the GRU cell are explored as an addition to our proposed architecture. Nearly all the fundamental ideas of both are the same. Both models can handle longer task sequences, and all include gated systems. In contrast to LSTM, GRU reduces the data flow by combining the forget gate and the input gate into one update gate. As a result, it has a simpler structure, fewer parameters, and a faster convergence speed.  
Each cell in the GRU computes its internal state (\text{$c_t$}) and hidden state (\text{$h_t$}) based on two gating units that regulate the flow of information: update gate (\text{$u_t$}) and reset gate (\text{$r_t$}). The update gate decides the amount of information the unit updates its content from the input features while the reset gate decides the amount of previous information to forget. W and b represent the weight and bias matrices in the training process. $\sigma(.)$ and tanh(.) denote the sigmoid and hyperbolic tangent activation functions, respectively. The computations are repeated for each element in the modeled sequence {\text($t$}).  The specific calculations can be expressed as:
\begin{equation}
	u_t = \sigma(W_u [gc(S_t, A), h_{t-1}]+b_u)
\end{equation}
\begin{equation}\label{eqdvff22}
	r_t = \sigma(W_r [gc(S_t, A), h_{t-1}]+b_r)
\end{equation}
\begin{equation}\label{edvqdvff22}
	c_t = tanh(W_c [gc(S_t, A), (r_t*h_{t-1})]+b_c)
\end{equation}
\begin{equation}\label{edvff22}
	h_t = (1-u_t)*c_t + u_t * h_{t-1}
\end{equation}

The traffic loop detectors data and its corresponding adjacency matrix were used to train and evaluate the proposed GC-GRU architecture to predict the future Traffic Index on the General Purpose (GP) lane. The training set is from January 1st 2020 to May 31th 2020 with 15-minute time interval. In other words, the number of  rows in the provided data is 14,551 reporting speeds at different segments (87 detectors) and time steps. The respective adjacency matrix provided for the data had a few additional nodes so it was pruned by detector (node) number to match the information provided in the traffic data. A few ‘nones’ also existed in the data, indicating non-available information so we change them to ‘0’, since the model only accepts integers. 
The test for the model for the traffic4cast challenge is performed on a separately provided test data for 15 days; from June 1st 2020 to June 16th 2020 also with 15-minute time interval. Each day provides 36 time-steps at which the prediction is to be made for the future 12 time-steps. The ground truth is not provided so the predicted values are then submitted as a json file to the leaderboard for evaluation using the MAPE calculated as:
\begin{equation}
	MAPE = \frac{1}{n} \Sigma^{n}_{i=1} (\frac{yi - \hat{yi}}{yi}) 
\end{equation}
where {\text{$y_i$}} and {\text{$\hat{yi}$}} represent the ground truth and predicted value respectively. The training/testing split on the data was experimented with a split of 80/20 and 90/10. Five datasets are then input into the model; training data, training labels, test data, test labels and adjacency matrix. 

The training/testing split on the data was experimented with a split of 80/20 and 90/10. Five datasets are then input into the model; training data, training labels, test data, test labels and adjacency matrix. 

Once the training data was cleaned, pre-processed and split for the model, it is fed into the GC cell encoder as a normalized tensor of shape (\text{$N, T_{in}, D$}) where N is the number of rows used for training, Tin is the length of time-steps and D is the number of detectors. The outputs from the GC network \text{$gc(S_t,A)$ }for each time is then fed into the encoder GRU cell to encode the temporal features between each time index. The decoder is then used to shape the output in the desired tensor of shape (\text{$N, T_{out}, D$}) where Tout is the length of time-steps for the forecasted horizon. The decoder GC cell uses the encoded intermediate representation vector to decode the spatial relationships and then passes it to the decoder GRU cell to decode the temporal relationships between output time-steps. The flattened output is then passed into a fully connected network layer which uses multi-output sigmoid activated layers to generate the predicted tensor which was then denormalized and reshaped to match the submission format.   
The loss function used for the proposed architecture is \text{$L2$}, defined as the sum of all the squared differences between the ground truth and predicted values then adding a ‘regularization term’ to avoid overfitting with a lambda loss of 0.0015 applied on the weights of the respective prediction ($i$). The loss function can be expressed as: 

\begin{equation}
		L(Y_{true}, Y_{pred}) = \frac{1}{N x T_{out} x D} \Sigma^{N}_{n} \Sigma^{T_{out}}_{t} 		
		\Sigma^{D}_{d} \mid ({T_{true}})_{n,t,d} - ({T_{pred}})_{n,t,d} \mid^2 + \lambda \Sigma^{M}_{i=1} \mid w_i \mid
\end{equation}

\subsection{Transformer Model for Traffic Forecasting}

%\begin{figure}[b!]
%	\centering
%	{\includegraphics[scale=1.2]{maged/Picture20}}
%	\caption{Transformer Model Architecture}
%	\label{fig_20}
%\end{figure}
The transformer architecture was originally designed for natural language processing (NLP) to provide contextual meaning for word tokens, which was a missing concept in feedforward networks like RNN, LSTM, and GRU. Prior to the introduction of transformers in NLP, word tokens were typically passed sequentially through NLP architectures. This method results in local understanding of word tokens rather than global understanding. Transformers, on the other hand, introduced a self-attention mechanism that allows each word token to attend to all other word tokens at the same time, giving it a global contextual meaning. Currently, transformers remain the state-of-art model for NLPs and other research areas such as computer vision.

%\begin{table}[h!]
%	\centering
%	\caption{Transformer trained model Parameters}
%	\label{table_5}
%	\resizebox{6cm}{!}{%
	%	\begin{tabular}{l l l}
		%	\hline
		%	\textbf{Model Parameter} & \textbf{Value}  \\ \hline \hline
		%	Batch size  & 40 \\
		%	Sequence length  & 36 \\
%			Prediction length &  12 \\
%			Learning rate   & 0.00001 \\
%			Epochs  &  100 \\
%			Minimum delta  &  0.0005 \\
%			Patience  & 10 \\
%			Model dimension  & 512 \\
%			Model number of heads  &  8 \\
%			Model number of layers  & 6 \\
%			Model dropout  & 0.3 \\
%			
%			\hline
%	\end{tabular}}
%\end{table}

The transformer network employs an encoder-decoder architecture similar to that of recurrent neural networks. The difference is that the input sequence can be passed in parallel. The encoder block accepts both input embeddings and positional embeddings. The encoder network is made up of a multi-head attention and a feedforward neural network. The attention layer computes an attention vector for each word token. The attention vectors are fed into the feedforward network, one vector at a time. Each attention network is self-contained, allowing for parallelization. The feedforward network is a simple neural network that is applied to each of the attention vectors. In practice, feedforward networks are used to convert attention vectors into a format that can be processed by the next encoder or decoder block. The feedforward network's final output is passed to the decoder block. The decoder block is made up of three components, two of which are similar to the encoder block. The decoder output is passed through a linear layer before being passed through a SoftMax to calculate probabilities. The Transformer model used as a benchmark was adopted from [52].

%\begin{figure}[b!]
%	\centering
%	{\includegraphics[scale=0.99]{maged/Picture21}}
%	\caption{LSTM Model Architecture}
%	\label{fig_21}
%\end{figure}

\subsection{LSTM Model for sequence-to-sequence traffic forecasting}
LSTM, or long short-term memory, is a particular class of RNN that can learn long-term sequences. Long-term reliance issues are specifically avoided in its design. Its method of operation involves recalling lengthy sequences over an extended period of time. The fact that each LSTM cell has a mechanism involved contributes to the popularity of LSTM. In a typical RNN cell, the activation layer transforms the input at the time stamp and the hidden state from the previous time step into a new state. In contrast, the LSTM process is a little more complicated because it requires input from three different states at once: the current input state, the short-term memory from the previous cell, and finally the long-term memory.
An outside input is received by the input gate, which processes the fresh data. The forget gate chooses the ideal time lag for the input sequence by determining when to forget the prior state. The output gate produces output for the LSTM cell using all the calculated results. In language models, a soft-max layer is typically introduced to control the NN's final output. On the output layer of the LSTM cell in our traffic flow prediction model, a linear regression layer is used. The LSTM model used as a benchmark was adopted from [52].

%\begin{table}[h!]	
%	\centering	
%	\caption{LSTM trained model Parameters}	
%	\label{table_6}	
%	\resizebox{6cm}{!}{%		
%		\begin{tabular}{l l l}			
%			\hline			
%			\textbf{Model Parameter} & \textbf{Value}  \\ \hline \hline			
%			Batch size  & 40 \\			
%			Sequence length  & 36 \\			
%			Prediction length &  12 \\			
%			Learning rate   & 0.00001 \\			
%			Epochs  &  100 \\			
%			Minimum delta  &  0.0005 \\			
%			Patience  & 10 \\			
%			Input dimension  & 87 \\			
%			Hidden dimension  &  87 \\		
%			Output dimension  & 87 \\		
%			\hline			
%	\end{tabular}}	
%\end{table}
\section{Results}
Table 1 summarizes the training time, inference time and MAPE metrics for our proposed model along with the benchmark models. A historical average (HA) model was also added as an additional benchmark to evaluate the performance of a simple statistical estimation. Our model performance on the provided test data ranked second in terms of MAPE which is very close to Transformer’s performance. We also evaluated additional parameters such as training and inference time and found that LSTM had the fastest training time which was expected given the simplicity of the model architecture when compared to the other models. It’s worth noting that our model not only had the fastest inference time, which is one of the crucial factors when implementing the model in real-time applications, but also had a training time that is six times faster than transformer. Fig 3 reflects a violin plot of the distribution of errors (MAE) along different models with a scaled distribution amplitude. HA has the highest distribution of errors followed by LSTM. Comparing GC-GRU-N and Transformers at bigger errors, we can observe that GC-GRU-N has a smaller distribution. 

\begin{figure}[b!]
	\centering
	{\includegraphics[scale=0.80]{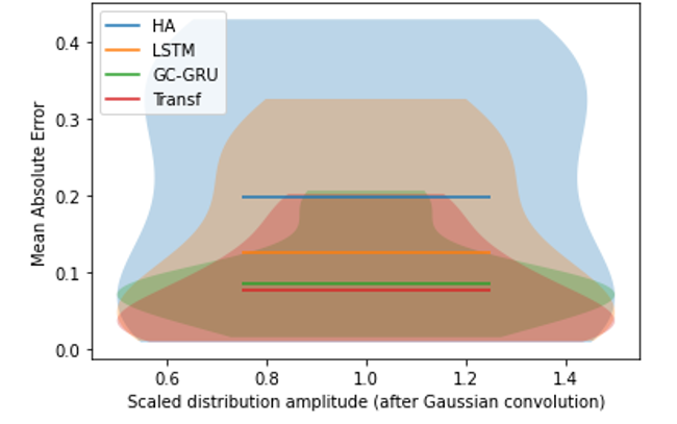}}
	\caption{MAE visualized distribution over segments}
	\label{fig_22}
\end{figure}

\begin{table}[h!]
	\centering
	\caption{Summary of model results}
	\label{table_77}
	\resizebox{9cm}{!}{%
		\begin{tabular}{l l l l}
			\hline
			\textbf & \textbf{LSTM} & \textbf{Transformer} & \textbf{GC-GRU-N}\\ \hline \hline
			 MAPE  & 4.50 & 3.12 & 3.16 \\
			Training time (sec)  & 76.23 & 1,397.88 & 217.79 \\
			Inference time (sec)  &  4.58 & 8.84 & 2.04 \\
			
			\hline
	\end{tabular}}
\end{table}

The proposed model was implemented with the aid of Tensorflow deep learning library [49]   and the GridSearchCV was used from scikit-learn [50] to find the optimal hyperparameters that can optimize the model performance. Table 5 summarizes the hyperparameters values used and Fig 6 illustrates the results in a 3D bubble plot along different model parameters, colored by MAPE. The selected bubble highlights the parameters for the best performing model.
\begin{figure}[b!]
	\centering
	{\includegraphics[scale=0.88]{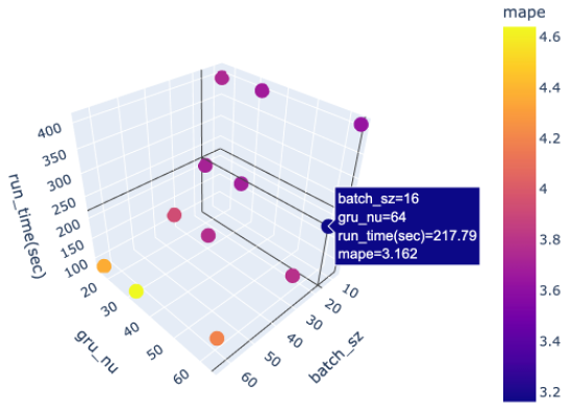}}
	\caption{3D bubble plot of performed experiments}
	\label{fig_23}
\end{figure}

%\begin{table}[h!]
%	\centering
%	\caption{GC-GRU Model training parameters}
%	\label{table_78}
%	\resizebox{7.5cm}{!}{%
%		\begin{tabular}{l l l}
%			\hline
%			\textbf{Model Parameter} & \textbf{Value}  \\ \hline \hline
%			Training ratio   & 0.8, 0.9  \\
%			Sequence length  & 36 \\
%			Prediction length &  12 \\
%			Learning rate   & 0.0001, 0.001, 0.01  \\
%			Batch Size   &  8, 16, 32, 64  \\
%			Epochs   &  20, 50  \\
%			Dropout  & 0.1, 0.2, 0.5 \\
%			GC layers   & 2, 3, 4  \\
%			GC layer size   &  4, 8, 12, 16  \\
%			GRU layers   & 2, 3, 4  \\
%			GRU layer size    & 16, 32, 64  \\
%			Optimizer    & AdamOptimizer  \\
			
%			\hline
%	\end{tabular}}
%\end{table}
\begin{table}[h!]
	\centering
	\caption{Prediction results of the proposed model and baseline models}
	\label{table_8}
	\resizebox{8.8cm}{!}{%
		\begin{tabular}{l l l l l l}
			\hline
			\textbf{Horizon} & \textbf{Metric} & \textbf{HA} & \textbf{LSTM} & \textbf{Transformer} & \textbf{GC-GRU-N}  \\ \hline \hline
			\multicolumn{6}{c}{$1-hour forecast$} \\
			& MAE  & 0.131 & 0.07 & 0.048 & \textbf{0.043} \\
			15 min & RMSE & 0.221 & 0.108 & 0.083 & \textbf{0.066} \\
			& MAPE & 0.248 & 0.109 & 0.076 & \textbf{0.069} \\
			
			& MAE  & 0.132 & 0.079  & 0.053  & \textbf{0.052} \\
			30 min & RMSE & 0.231  & 0.118  & 0.087  & \textbf{0.08}  \\
			& MAPE & 0.272  & 0.132  & \textbf{0.079} & 0.086   \\
			
			& MAE  & 0.141  & 0.08 & 0.058 & \textbf{0.054} \\
			45 min & RMSE & 0.239 & 0.122 & 0.092 & \textbf{0.082} \\
			& MAPE & 0.294  & 0.144  & \textbf{0.089}  & 0.092  \\
			
			& MAE  & 0.154  & 0.088  & 0.059  & \textbf{0.058}  \\
			60 min & RMSE & 0.252  & 0.136  & 0.094  & \textbf{0.089} \\
			& MAPE & 0.314  & 0.157  & \textbf{0.09}  & 0.095  \\	
			
			\multicolumn{6}{c}{$2-hour forecast$} \\
			& MAE  & 0.171  & 0.102  & \textbf{0.068}  & 0.075  \\
			75 min & RMSE & 0.276  & 0.16  & 0.11  & \textbf{0.105}  \\
			& MAPE & 0.381  & 0.207  & \textbf{0.115}  & 0.131  \\
			
			& MAE  & 0.162  & 0.1   & 0.07   & \textbf{0.067}  \\
			90 min & RMSE & 0.27   & 0.158   & 0.108   & \textbf{0.097}   \\
			& MAPE & 0.372   & 0.208   & \textbf{0.114}   & 0.118    \\
			
			& MAE  & 0.172   & 0.104  & 0.072  & \textbf{0.071}  \\
			105 min & RMSE & 0.277  & 0.168  & 0.105  & \textbf{0.100}  \\
			& MAPE & 0.404   & 0.236   & \textbf{0.127}  & 0.135   \\
			
			& MAE  & 0.181   & 0.111   & \textbf{0.079}   & 0.086   \\
			120 min & RMSE & 0.281   & 0.18   & \textbf{0.113}   & 0.114   \\
			& MAPE & 0.413   & 0.253   & \textbf{0.141}   & 0.159   \\	
			
			\multicolumn{6}{c}{$3-hour forecast$} \\
			& MAE  & 0.18   & 0.112   & \textbf{0.076}   & 0.084   \\
			135 min & RMSE & 0.275   & \textbf{0.182}   & 0.111   & 0.115   \\
			& MAPE & 0.393   & 0.251   & \textbf{0.139}   & 0.155   \\
			
			& MAE  & 0.176   & 0.12    & \textbf{0.08}    & 0.096   \\
			150 min & RMSE & 0.272    & \textbf{0.195}    & 0.116    & 0.131    \\
			& MAPE & 0.393    & 0.273    & \textbf{0.155}    & 0.183     \\
			
			& MAE  & 0.176    & 0.126   & \textbf{0.079}   & 0.1   \\
			165 min & RMSE & 0.274   & 0.205   & \textbf{0.121}   & 0.145   \\
			& MAPE & 0.39    & 0.285    & \textbf{0.157}   & 0.197    \\
			
			& MAE  & 0.158    & 0.121    & \textbf{0.071}    & 0.095    \\
			180 min & RMSE & 0.257    & 0.196    & \textbf{0.114}    & 0.146    \\
			& MAPE & 0.347    & 0.265    & \textbf{0.139}   & 0.19    \\
			
			\hline
	\end{tabular}}
\end{table}

Table 6 presents the results for each of the trained models. The models are evaluated  along the forecasted horizon of 12 time-steps or 3-hours (12 time-steps divided by four 15-minute intervals per hour). The evaluation is performed on the test data split from the provided traffic features data and all three models use the same exact data for testing. We split the table along each forecasted hour to better understand how each model performs on different horizons. The metrics used for evaluation are Mean Absolute Error (MAE), Root Mean Squared Error (RMSE) and Mean Absolute Percentage Error (MAPE). MAE is defined as the average magnitude of the differences between the predicted and observed true values while, RMSE is the standard deviation or a measure of how spread the differences between predictions and observed true values are, and MAPE is defined as how far the predicted values are away from the corresponding observed truth on average. 

%\lipsum

%\lipsum
In general, the proposed GC-GRU-N model achieved the best results along 1-hour forecasted horizon while transformers were dominant along the 3-hour forecasted horizon. Along the 2-hours horizon both our model and Transformer seemed to share equal success when looking at the performance metrics. The HA model performed the worst in predictions demonstrating the need for a deep-learning model. The stand-alone LSTM model seemed to struggle in performing predictions with errors almost double in magnitude when compared to the other two models. This signifies the fact that learning from temporal dependency only is not enough and the synergetic effect of combining spatial and temporal dependencies is critical to maximize the network’s learning capabilities for more accurate predictions. 

\begin{figure*}[b!]
	\centering
	\includegraphics[scale=0.34]{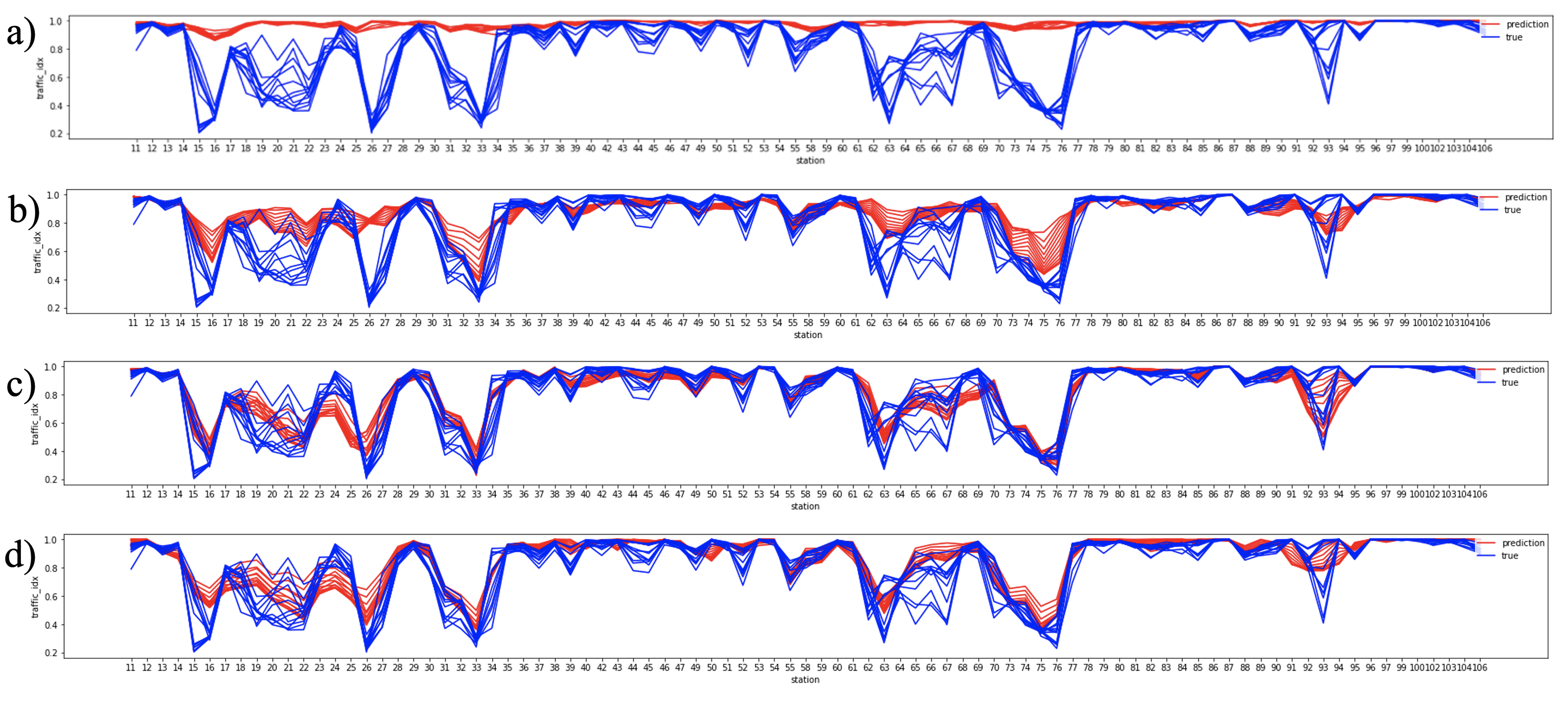}
	\caption{Visualization results of all prediction horizons along traffic detectors for: (a) HA, (b) LSTM,  (c) Transformer, (d) GC-GRU-N }
	\label{fig_24}
\end{figure*}
\begin{figure*}[b!]
	\centering
	{\includegraphics[scale=0.40]{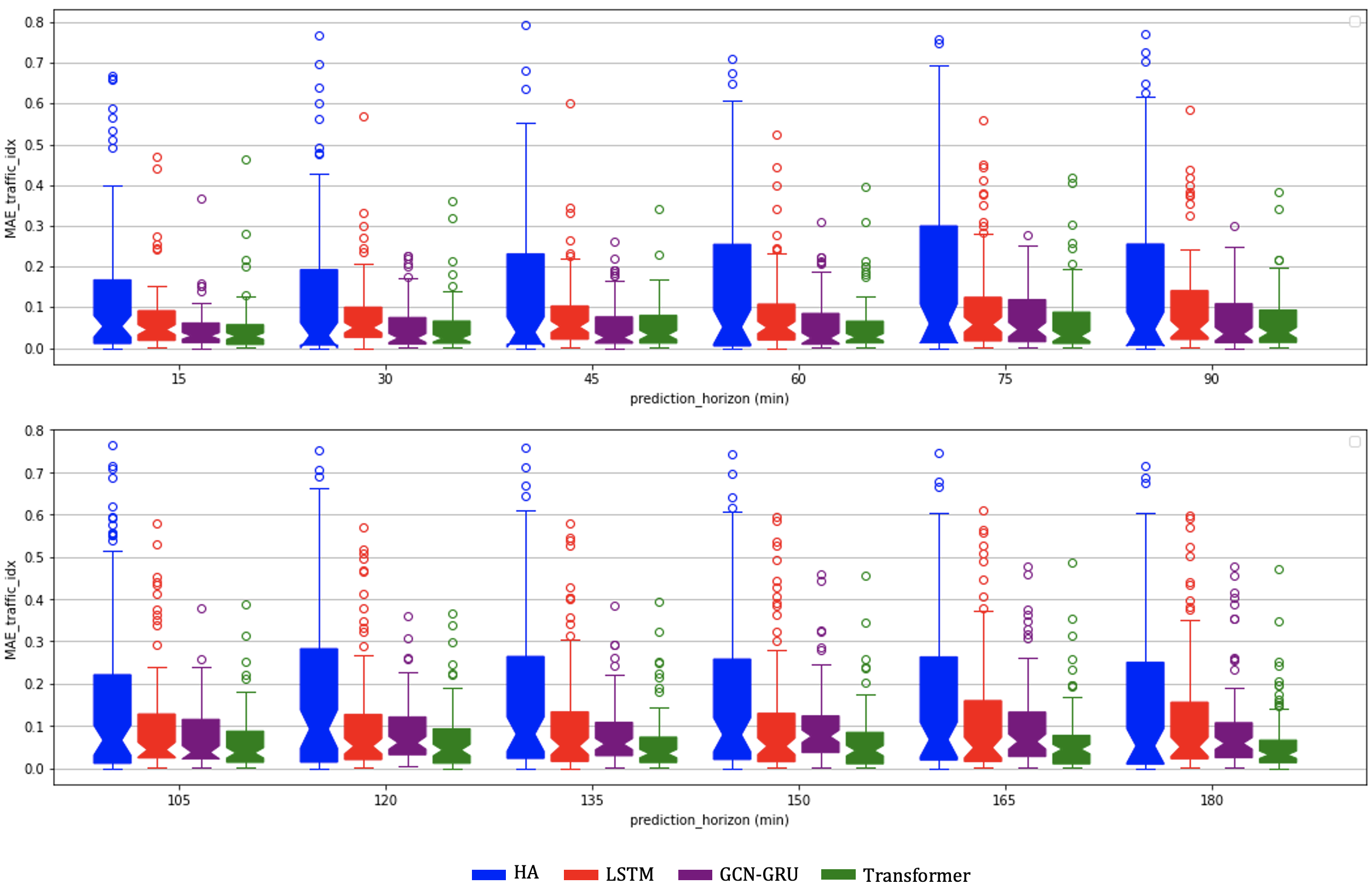}}
	\caption{MAE box-plots of all traffic stations results along future prediction horizons}
	\label{fig_25}
\end{figure*}

\subsection{Spatial Analysis of trained models}
Fig 7 presents the predicted values of each model and the observed truth along each detector. The plot is for all predicted time-steps so we have twelve lines for predictions and twelve for the true values. Once again, the HA had the worst performance with predictions very far-off the ground truth. Generally, we can observe that Transformer and GC-GRU-N predicted peak values much better than LSTM. Transformer predictions seem to be much closer to true values during peaks than GC-GRU-N and we suspect that may be due to the smooth filter applied by the GC that captures spatial features by constantly moving the filter. This leads to much smoother peaks. The results also show that there still seems to be a certain error even at non-peaks which can be explained by the fact that there are times when we don’t have traffic data or the values are very small.  

%\lipsum

%\lipsum

\subsection{Temporal Analysis of trained models}
Fig 8 presents the MAE box-plot for each model along the forecasted horizon. Box plots provide a standardized way of interpreting he distribution of errors based on the minimum, maximum, median, 25th and 75th percentiles and the outliers. In a box-plot, density of values is inversely proportional to the size of the box which means that smaller boxes have higher number of values packed in their respective range. HA model had the largest amount to errors as expected. MAE box-plots for the other models increase over the length of the forecasted horizon indicating a positive association between the forecasting errors and the distance into future predictions. LSTM seemed to have the highest range and distribution of errors when compared to Transformer and GC-GRU-N. Comparing our proposed model to Transformer, we can notice that our model had smaller errors (better performance) on the closer predictions up to 60 minutes ahead, after which the Transformer model has a better performance on the longer horizon.

\section{Conclusion}
In this paper, we introduced a GC-GRU-N based neural network-based traffic forecasting method. To model the Seattle loop detector data, we utilize a graph convolution, in which the nodes on the graph represent the roads, the edges indicate the connections between the roads, and the attribute of the nodes on the graph is the traffic information on the roads. In order to acquire the spatial dependence, the GC cell is used to capture the topological structure of the graph. In order to obtain the temporal dependence, the GRU cell is used to capture the dynamic change in node attribute. We then performed a comparative analysis on  the proposed model along with benchmark models such as: HA, LSTM and Transformer, on test data also provided by [54]. In summary, our GC-GRU-N model performance on the provided test data ranked second with a MAPE of 3.16 which is very close to Transformer’s performance of 3.12. In terms of training and inference time, we found that LSTM had the fastest training time which was expected given the simplicity of the model architecture when compared to the other models. It’s worth noting that our model not only had the fastest inference time, but also had a training time that is six times faster than Transformer. Comparative analysis of the results on the test data demonstrates that the proposed GC-GRU-N is a strong competitor to state-of-the-art traffic forecasting approaches.

\end{document}